\documentclass{article} % For LaTeX2e
\usepackage{iclr2017_conference,times}
\usepackage{hyperref}
\usepackage{url}

\usepackage{times}
\usepackage{url}
\usepackage{latexsym}
\usepackage{graphicx}
\usepackage{amsmath}
\usepackage{amsfonts}
\usepackage{amssymb}
\usepackage{dsfont}
\usepackage[english]{babel}
\usepackage{color}
\usepackage{booktabs}
\usepackage{multicol}
\usepackage{multirow}

\title{A Compare-Aggregate Model for Matching Text Sequences}
\author{Shuohang Wang\\
		School of Information Systems\\
		Singapore Management University\\ 
		\texttt{shwang.2014@phdis.smu.edu.sg}
	\And 
	Jing Jiang\\
	School of Information Systems\\
	Singapore Management University\\
	\texttt{jingjiang@smu.edu.sg}}
\def\ignore#1{}

\begin{document}

\maketitle
\begin{abstract}
Many NLP tasks including machine comprehension, answer selection and text entailment require the comparison between sequences. 
Matching the important units between sequences is a key to solve these problems. 
In this paper, we present a general ``compare-aggregate'' framework that performs word-level matching followed by aggregation using Convolutional Neural Networks.
We particularly focus on the different comparison functions we can use to match two vectors.
% will still start from word-level matching and these matching representations will be constructed to sequence-level matching through Convolutional Neural Networks. Besides, we have a further analysis on 8 different methods for the word-level matching through 4 different data sets in different tasks. 
We use four different datasets to evaluate the model.
We find that some simple comparison functions based on element-wise operations can work better than standard neural network and neural tensor network. 
% Overall, Finally, our model combined with a suitable word-level matching method can outperform most of the previous methods.
\end{abstract}

\section{Introduction}
\label{sec:intro2}

Many natural language processing problems involve matching two or more sequences to make a decision.
For example, in textual entailment, one needs to determine whether a hypothesis sentence can be inferred from a premise sentence~\citep{bowman2015large:EMNLP}.
In machine comprehension, given a passage, a question needs to be matched against it in order to find the correct answer~\citep{richardsonmctest:EMNLP2013,MovieQA:cvpr2016}.
Table~\ref{sample} gives two example sequence matching problems.
In the first example, a passage, a question and four candidate answers are given.
We can see that to get the correct answer, we need to match the question against the passage and identify the last sentence to be the answer-bearing sentence.
In the second example, given a question and a set of candidate answers, we need to find the answer that best matches the question.
Because of the fundamental importance of comparing two sequences of text to judge their semantic similarity or relatedness, sequence matching has been well studied in natural language processing.

With recent advances of neural network models in natural language processing, a standard practice for sequence modeling now is to encode a sequence of text as an embedding vector using models such as RNN and CNN.
To match two sequences, a straightforward approach is to encode each sequence as a vector and then to combine the two vectors to make a decision~\citep{bowman2015large:EMNLP,feng2015applying}.
However, it has been found that using a single vector to encode an entire sequence is not sufficient to capture all the important information from the sequence, and therefore advanced techniques such as attention mechanisms and memory networks have been applied to sequence matching problems~\citep{hermann:nips2015,hill:ICLR2016,rock:ICLR2016}.

A common trait of a number of these recent studies on sequence matching problems is the use of a ``compare-aggregate'' framework~\citep{wang:NAACL2016, he:naacl16, parikh:emnlp2016}.
In such a framework, comparison of two sequences is not done by comparing two vectors each representing an entire sequence.
Instead, these models first compare vector representations of smaller units such as words from these sequences and then aggregate these comparison results to make the final decision.
For example, the match-LSTM model proposed by \cite{wang:NAACL2016} for textual entailment first compares each word in the hypothesis with an attention-weighted version of the premise.
The comparison results are then aggregated through an LSTM.
\cite{he:naacl16} proposed a pairwise word interaction model that first takes each pair of words from two sequences and applies a comparison unit on the two words.
It then combines the results of these word interactions using a similarity focus layer followed by a multi-layer CNN.
\cite{parikh:emnlp2016} proposed a decomposable attention model for textual entailment, in which words from each sequence are compared with an attention-weighted version of the other sequence to produce a series of comparison vectors.
The comparison vectors are then aggregated and fed into a feed forward network for final classification.

Although these studies have shown the effectiveness of such a ``compare-aggregate'' framework for sequence matching, there are at least two limitations with these previous studies:
(1) Each of the models proposed in these studies is tested on one or two tasks only, but we hypothesize that this general framework is effective on many sequence matching problems. 
There has not been any study that empirically verifies this.
(2) More importantly, these studies did not pay much attention to the comparison function that is used to compare two small textual units.
Usually a standard feedforward network is used~\citep{hu2014convolutional,wang:NAACL2016} to combine two vectors representing two units that need to be compared, e.g., two words.
However, based on the nature of these sequence matching problems, we essentially need to measure how semantically similar the two sequences are.
Presumably, this property of these sequence matching problems should guide us in choosing more appropriate comparison functions.
Indeed \cite{he:naacl16} used cosine similarity, Euclidean distance and dot product to define the comparison function, which seem to be better justifiable.
But they did not systematically evaluate these similarity or distance functions or compare them with a standard feedforward network.

In this paper, we argue that the general ``compare-aggregate'' framework is effective for a wide range of sequence matching problems.
We present a model that follows this general framework and test it on four different datasets, namely, MovieQA, InsuranceQA, WikiQA and SNLI.
The first three datasets are for Question Answering, but the setups of the tasks are quite different.
The last dataset is for textual entailment.
More importantly, we systematically present and test six different comparison functions.
We find that overall a comparison function based on element-wise subtraction and multiplication works the best on the four datasets.

The contributions of this work are twofold:
(1) Using four different datasets, we show that our model following the ``compare-aggregate'' framework is very effective when compared with the state-of-the-art performance on these datasets.
(2) We conduct systematic evaluation of different comparison functions and show that a comparison function based on element-wise operations, which is not widely used for word-level matching, works the best across the different datasets.
We believe that these findings will be useful for future research on sequence matching problems.
We have also made our code available online.\footnote{\url{https://github.com/shuohangwang/SeqMatchSeq}}

\begin{table}[]
\small
\label{sample}
\centering
\begin{tabular}{ll}
\begin{tabular}{l}
\toprule
\multicolumn{1}{p{6cm}}{
\textbf{Plot}: ... Aragorn is crowned King of Gondor and taking Arwen as his queen before all present at his coronation bowing before Frodo and the other Hobbits . The Hobbits return to \textbf{the Shire} where Sam marries Rosie Cotton . ...} \\ \midrule
\multicolumn{1}{p{6cm}}{\textbf{Qustion}: Where does Sam marry Rosie? } \\ \midrule
\multicolumn{1}{p{6cm}}{\textbf{Candidate answers}: 0) Grey Havens. 1) Gondor. \textbf{2) The Shire}. 3) Erebor. 4) Mordor. }\\
\bottomrule
\end{tabular}
&
\begin{tabular}{l}
\toprule
\multicolumn{1}{p{6cm}}{\textbf{Question}: can i have auto insurance without a car}                                                                                                                                                                                                                                                                                                                                            \\ \midrule
\multicolumn{1}{p{6cm}}{
\textbf{Ground-truth answer}: yes, it be possible have auto insurance without own a vehicle. you will purchase what be call a name ...
% nonowner policy...
} \\ \midrule
\multicolumn{1}{p{6cm}}{
\textbf{Another candidate answer}: insurance not be a tax or merely a legal obligation because auto insurance follow a car...
% you shall have auto insurance ...
%any time you own a car. if you decide register the car, most state will require insurance covering liability for the operation of the car...
% if you be go store a car you shall consider comprehensive auto insurance protect your car from a wide range of peril.
}\\ \bottomrule
\end{tabular}
\end{tabular}
\normalsize
\caption{The example on the left is a machine comprehension problem from MovieQA, where the correct answer here is \textbf{The Shire}. 
The example on the right is an answer selection problem from InsuranceQA.}
\end{table}

\section{Method}
\begin{figure}[]
\centering
\includegraphics[width=5.5in]{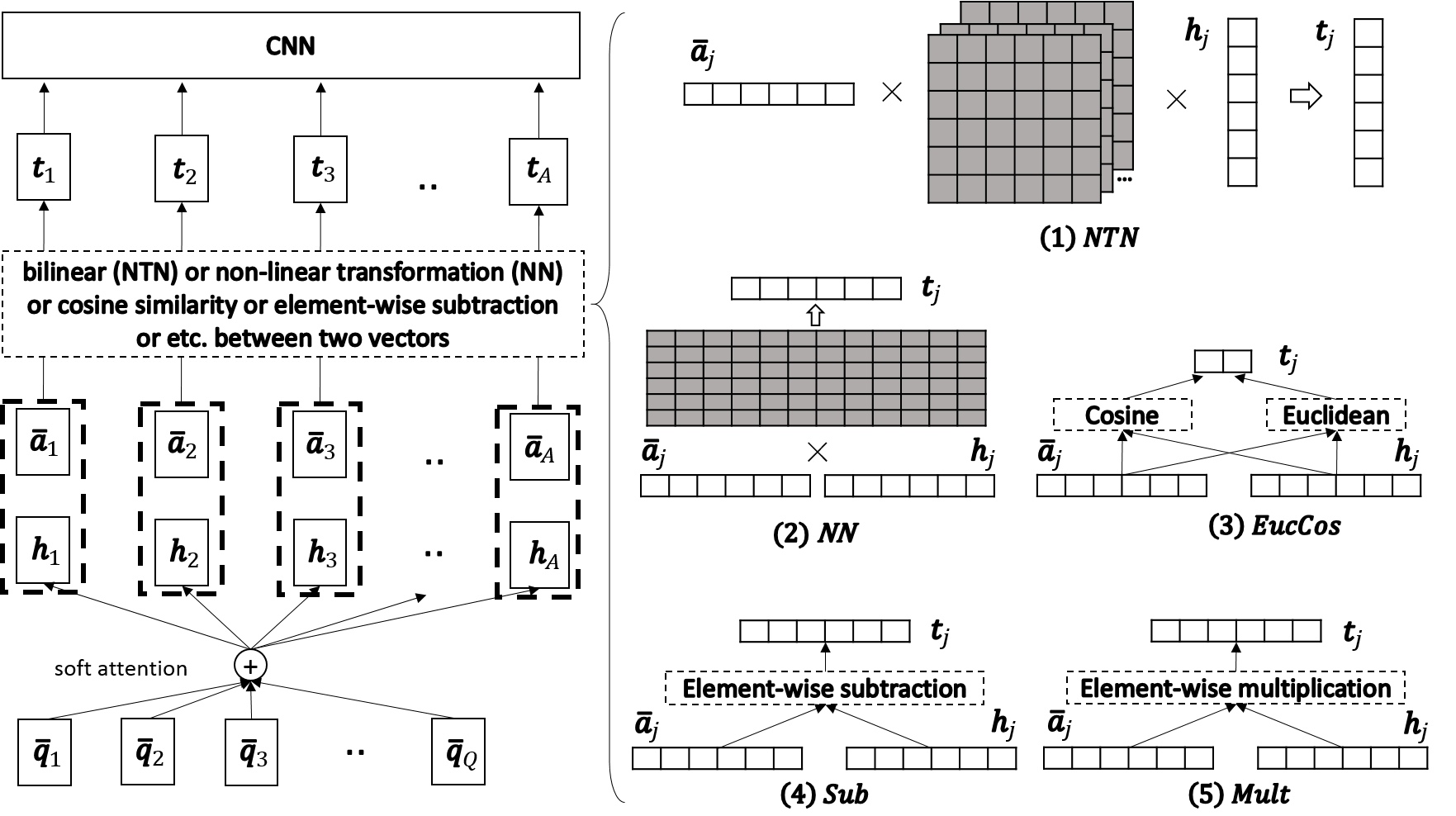}
\caption{The left hand side is an overview of the model.
	The right hand side shows the details about the different comparison functions. 
	The rectangles in dark represent parameters to be learned. 
%$\odot$ represents the element-wise multiplication and 
$\times$ represents matrix multiplication.}
\label{fig:model}
\end{figure}
In this section, we propose a general model following the ``compare-aggregate'' framework for matching two sequences.
This general model can be applied to different tasks.
We focus our discussion on six different comparison functions that can be plugged into this general ``compare-aggregate'' model.
In particular, we hypothesize that two comparison functions based on element-wise operations, \textsc{Sub} and \textsc{Mult}, are good middle ground between highly flexible functions using standard neural network models and highly restrictive functions based on cosine similarity and/or Euclidean distance.
As we will show in the experiment section, these comparison functions based on element-wise operations can indeed perform very well on a number of sequence matching problems.

% In this section, we will present a general model for matching two sequences which can be extended to different tasks. We will also show 8 different word level matching representations used in the sequence matching. An overview of our methods is shown in Figure~\ref{fig:model}. 

%In all these tasks, we use matrix $\mathbf{Q} \in \mathbb{R}^{d \times Q}$ to represent the question/premise, matrix $\mathbf{A}_k \in \mathbb{R}^{d \times A_k}$ to represent the sequence of answer/hypothesis, matrix $\mathbf{P} \in \mathbb{R}^{d \times A}$ to represent the plot specifically in MovieQA task, where $Q$ is the length of the question/premise, $A_k$ the length of the $k^\text{th}$ answer, $P$ the length of the plot. 

\subsection{Problem Definition and Model Overview}

The general setup of the sequence matching problem we consider is the following.
We assume there are two sequences to be matched.
We use two matrices $\mathbf{Q} \in \mathbb{R}^{d \times Q}$ and $\mathbf{A} \in \mathbb{R}^{d \times A}$ to represent the word embeddings of the two sequences, where $Q$ and $A$ are the lengths of the two sequences, respectively, and $d$ is the dimensionality of the word embeddings.
In other words, each column vector of $\mathbf{Q}$ or $\mathbf{A}$ is an embedding vector representing a single word.
Given a pair of $\mathbf{Q}$ and $\mathbf{A}$, the goal is to predict a label $y$.
For example, in textual entailment, $\mathbf{Q}$ may represent a premise and $\mathbf{A}$ a hypothesis, and $y$ indicates whether $\mathbf{Q}$ entails $\mathbf{A}$ or contradicts $\mathbf{A}$.
In question answering, $\mathbf{Q}$ may be a question and $\mathbf{A}$ a candidate answer, and $y$ indicates whether $\mathbf{A}$ is the correct answer to $\mathbf{Q}$.

We treat the problem as a supervised learning task.
We assume that a set of training examples in the form of $(\mathbf{Q}, \mathbf{A}, y)$ is given and we aim to learn a model that maps any pair of $(\mathbf{Q}, \mathbf{A})$ to a $y$.

% Given two different sequences, we use matrix $\mathbf{Q} \in \mathbb{R}^{d \times Q}$ to represent the word embeddings in one sequence, and matrix $\mathbf{A} \in \mathbb{R}^{d \times A}$ to represent the other, where $Q$ and $A$ represent the lengths of two sequences.

An overview of our model is shown in Figure~\ref{fig:model}.
The model can be divided into the following four layers:
\begin{enumerate}
	\item \textbf{Preprocessing:} We use a preprocessing layer (not shown in the figure) to process $\mathbf{Q}$ and $\mathbf{A}$ to obtain two new matrices $\overline{\mathbf{Q}} \in \mathbb{R}^{l \times Q}$ and $\overline{\mathbf{A}} \in \mathbb{R}^{l \times A}$.
	The purpose is to obtain a new embedding vector for each word in each sequence that captures some contextual information in addition to the word itself.
	For example, $\overline{\mathbf{q}}_i \in \mathbb{R}^l$, which is the $i^\text{th}$ column vector of  $\overline{\mathbf{Q}}$, encodes the $i^\text{th}$ word in $\mathbf{Q}$ together with its context in $\mathbf{Q}$.
	
	\item \textbf{Attention:} We apply a standard attention mechanism on $\overline{\mathbf{Q}}$ and $\overline{\mathbf{A}}$ to obtain attention weights over the column vectors in $\overline{\mathbf{Q}}$ for each column vector in $\overline{\mathbf{A}}$.
	With these attention weights, for each column vector $\overline{\mathbf{a}}_j$ in $\overline{\mathbf{A}}$, we obtain a corresponding vector $\mathbf{h}_j$, which is an attention-weighted sum of the column vectors of $\overline{\mathbf{Q}}$.
	
	\item \textbf{Comparison:} We use a comparison function $f$ to combine each pair of $\overline{\mathbf{a}}_j$ and $\mathbf{h}_j$ into a vector $\mathbf{t}_j$.
	
	\item \textbf{Aggregation:} We use a CNN layer to aggregate the sequence of vectors $\mathbf{t}_j$ for the final classification.
	
\end{enumerate}
 
Although this model follows more or less the same framework as the model proposed by \cite{parikh:emnlp2016}, our work has some notable differences.
First, we will pay much attention to the comparison function $f$ and compare a number of options, including a some uncommon ones based on element-wise operations.
Second, we apply our model to four different datasets representing four different tasks to evaluate its general effectiveness for sequence matching problems.
There are also some other differences from the work by \cite{parikh:emnlp2016}. 
For example, we use a CNN layer instead of summation and concatenation for aggregation.
Our attention mechanism is one-directional instead of two-directional.
% We find one-directional attention to be critical because for some tasks one of the sequences may be very long, making it computationally too expensive to apply two-directional attention.
% \jjcomment{Is the previous sentence correct?}

In the rest of this section we will present the model in detail. 
We will focus mostly on the comparison functions we consider.

\subsection{Preprocessing and Attention}

Our preprocessing layer uses a recurrent neural network to process the two sequences.
We use a modified version of LSTM/GRU in which we keep only the input gates for remembering meaningful words:
\begin{eqnarray}
\nonumber
\overline{\mathbf{Q}} & = & \sigma(\mathbf{W}^\text{i} \mathbf{Q} + \mathbf{b}^{\text{i}} \otimes \mathbf{e}_Q) \odot \tanh(\mathbf{W}^{\text{u}}\mathbf{Q}+\mathbf{b}^{\text{u}}\otimes \mathbf{e}_Q), \\
\overline{\mathbf{A}} & = & \sigma(\mathbf{W}^\text{i} \mathbf{A} + \mathbf{b}^{\text{i}} \otimes \mathbf{e}_A) \odot \tanh(\mathbf{W}^{\text{u}}\mathbf{A}+\mathbf{b}^{\text{u}}\otimes \mathbf{e}_A),
\end{eqnarray}
where $\odot$ is element-wise multiplication, and $\mathbf{W}^\text{i}, \mathbf{W}^\text{u}\in \mathbb{R}^{l\times d}$ and $\mathbf{b}^\text{i},\mathbf{b}^\text{u}\in \mathbb{R}^{l}$ are parameters to be learned.
The outer product $(\cdot \otimes \mathbf{e}_X)$ produces a matrix or row vector by repeating the vector or scalar on the left for $X$ times.

The attention layer is built on top of the resulting $\overline{\mathbf{Q}}$ and $\overline{\mathbf{A}}$ as follows:
\begin{eqnarray}
\nonumber
\mathbf{G} & = & \text{softmax} \left(   ( \mathbf{W}^{\text{g}} \overline{\mathbf{Q}}  + \mathbf{b}^{\text{g}} \otimes \mathbf{e}_Q)^{\text{T}} \overline{\mathbf{A}} \right), \\
\label{eqn:alpha}
\mathbf{H} & = & \overline{\mathbf{Q}} \mathbf{G},
\end{eqnarray}
where $\mathbf{W}^{\text{g}} \in \mathbb{R}^{l\times l}$ and $\mathbf{b}^{\text{g}} \in \mathbb{R}^{l}$ are parameters to be learned, $\mathbf{G}\in \mathbb{R}^{Q\times A}$ is the attention weight matrix, and $\mathbf{H} \in \mathbb{R}^{l\times A}$ are the attention-weighted vectors. 
Specifically, $\mathbf{h}_j$, which is the $j^\text{th}$ column vector of $\mathbf{H}$, is a weighted sum of the column vectors of $\overline{\mathbf{Q}}$ and represents the part of $\mathbf{Q}$ that best matches the $j^\text{th}$ word in $\mathbf{A}$.
Next we will combine $\mathbf{h}_j$ and $\overline{\mathbf{a}}_j$ using a comparison function.
% Here, each column of $\mathbf{H}$ can be treated as the word representation in $\mathbf{Q}$ that is more likely to be aligned with the corresponding column in $\mathbf{H}^\text{a}$. Next, we will make use of these aligned information and build the matched representation between them.

\subsection{Comparison}

The goal of the comparison layer is to match each $\overline{\mathbf{a}}_j$, which represents the $j^\text{th}$ word and its context in $\mathbf{A}$, with $\mathbf{h}_j$, which represents a weighted version of $\mathbf{Q}$ that best matches $\overline{\mathbf{a}}_j$.
Let $f$ denote a comparison function that transforms $\overline{\mathbf{a}}_j$ and $\mathbf{h}_j$ into a vector $\mathbf{t}_j$ to represent the comparison result.

A natural choice of $f$ is a standard neural network layer that consists of a linear transformation followed by a non-linear activation function.
For example, we can consider the following choice:
\begin{eqnarray}
\textsc{NeuralNet (NN):}& & \mathbf{t}_j = f(\overline{\mathbf{a}}_j, \mathbf{h}_j) = \text{ReLU}(\mathbf{W} \begin{bmatrix} \overline{\mathbf{a}}_j \\ \mathbf{h}_j \end{bmatrix}  + \mathbf{b}),
\end{eqnarray}
where matrix $\mathbf{W} \in \mathbb{R}^{l\times 2l}$ and vector $\mathbf{b}\in \mathbb{R}^{l}$ are parameters to be learned.
%\jjcomment{I suggest that we remove the superscript $\text{r}$ in $\mathbf{W}^\text{r}$ and $\mathbf{b}^\text{r}$. I think re-using the same symbol $\mathbf{W}$, $\mathbf{b}$, etc. for different comparison functions should not cause any confusion because each time we only use a single comparison function, i.e., we will never use multiple comparison functions together.}

Alternatively, another natural choice is a neural tensor network~\citep{socher2013:emnlp} as follows:
\begin{eqnarray}
\textsc{NeuralTensorNet (NTN):} & & \mathbf{t}_j = f(\overline{\mathbf{a}}_j, \mathbf{h}_j) = \text{ReLU}(\overline{\mathbf{a}}_j^\text{T} \mathbf{T}^{[1 \ldots l]} \mathbf{h}_j + \mathbf{b}),
\label{eqn:bilinear}
\end{eqnarray}
where tensor $\mathbf{T}^{[1 \ldots l]}\in \mathbb{R}^{l\times l\times l}$ and vector $\mathbf{b} \in \mathbb{R}^l$ are parameters to be learned.

However, we note that for many sequence matching problems, we intend to measure the semantic similarity or relatedness of the two sequences.
So at the word level, we also intend to check how similar or related $\overline{\mathbf{a}}_j$ is to $\mathbf{h}_j$.
For this reason, a more natural choice used in some previous work~\cite{} is Euclidean distance or cosine similarity between $\overline{\mathbf{a}}_j$ and $\mathbf{h}_j$.
We therefore consider the following definition of $f$:
\begin{eqnarray}
\textsc{Euclidean+Cosine (EucCos):} & & \mathbf{t}_j = f(\overline{\mathbf{a}}_j, \mathbf{h}_j) = \begin{bmatrix} \Vert \overline{\mathbf{a}}_j - \mathbf{h}_j \Vert _{2} \\ \cos(\overline{\mathbf{a}}_j, \mathbf{h}_j)\end{bmatrix}.
\label{eqn:cos}
\end{eqnarray}

Note that with \textsc{EucCos}, the resulting vector $\mathbf{t}_j$ is only a 2-dimensional vector.
Although \textsc{EucCos} is a well-justified comparison function, we suspect that it may lose some useful information from the original vectors $\overline{\mathbf{a}}_j$ and $\mathbf{h}_j$.
On the other hand, \textsc{NN} and \textsc{NTN} are too general and thus do not capture the intuition that we care mostly about the similarity between $\overline{\mathbf{a}}_j$ and $\mathbf{h}_j$.

To use something that is a good compromise between the two extreme cases, we consider the following two new comparison functions, which operate on the two vectors in an element-wise manner.
These functions have been used previously by \cite{tai2015improved:acl}.
\begin{eqnarray}
\textsc{Subtraction (Sub):} & & \mathbf{t}_j = f(\overline{\mathbf{a}}_j, \mathbf{h}_j) =  (\overline{\mathbf{a}}_j - \mathbf{h}_j) \odot (\overline{\mathbf{a}}_j - \mathbf{h}_j), \\
\textsc{Multiplication (Mult):} & & \mathbf{t}_j = f(\overline{\mathbf{a}}_j, \mathbf{h}_j) =  \overline{\mathbf{a}}_j \odot \mathbf{h}_j.
\end{eqnarray}
Note that the operator $\odot$ is element-wise multiplication.
For both comparison functions, the resulting vector $\mathbf{t}_j$ has the same dimensionality as $\overline{\mathbf{a}}_j$ and $\mathbf{h}_j$.

We can see that \textsc{Sub} is closely related to Euclidean distance in that Euclidean distance is the sum of all the entries of the vector $\mathbf{t}_j$ produced by \textsc{Sub}.
But by not summing up these entries, \textsc{Sub} preserves some information about the different dimensions of the original two vectors.
Similarly, \textsc{Mult} is closely related to cosine similarity but preserves some information about the original two vectors.

%We further note that \textsc{Sub} and \textsc{Mult} can be generalized to a weighted version as follows:
%\begin{eqnarray}
%\textsc{Weighted Sub (WeightedSub):} & &  \\
%\textsc{Weighted Mult (WeightedMult):} & & 
%\end{eqnarray}

Finally, we consider combining \textsc{Sub} and \textsc{Mult} followed by an NN layer as follows:
\begin{eqnarray}
\textsc{SubMult+NN:} & & \mathbf{t}_j = f(\overline{\mathbf{a}}_j, \mathbf{h}_j) =  \text{ReLU}(\mathbf{W} \begin{bmatrix} (\overline{\mathbf{a}}_j - \mathbf{h}_j) \odot (\overline{\mathbf{a}}_j - \mathbf{h}_j) \\ \overline{\mathbf{a}}_j \odot \mathbf{h}_j \end{bmatrix}  + \mathbf{b}).
\end{eqnarray}
%\jjcomment{Similarly, I suggest that we remove the superscript $\text{r}$ for the parameters.}

In summary, we consider six different comparison functions: \textsc{NN}, \textsc{NTN}, \textsc{EucCos}, \textsc{Sub}, \textsc{Mult} and \textsc{SubMult+NN}.
Among these functions, the last three (\textsc{Sub}, \textsc{Mult} and \textsc{SubMult+NN}) have not been widely used in previous work for word-level matching.

\subsection{Aggregation}

After we apply the comparison function to each pair of $\overline{\mathbf{a}}_j$ and $\mathbf{h}_j$ to obtain a series of vectors $\mathbf{t}_j$, finally we aggregate these vectors using a one-layer CNN~\citep{kim:emnlp14}:
\begin{eqnarray}
\mathbf{r} & = & \text{CNN}([\mathbf{t}_1, \ldots, \mathbf{t}_A]).
\label{eqn:aggregate}
\end{eqnarray}
$\mathbf{r}\in \mathbb{R}^{nl}$ is then used for the final classification, where $n$ is the number of windows in CNN.

\section{Experiments}

\begin{table}[]
\centering
\footnotesize
\begin{tabular}{ccccccccccccc}
\toprule
\multirow{2}{*}{}                              & \multicolumn{3}{c}{MovieQA} & \multicolumn{3}{c}{InsuranceQA} & \multicolumn{3}{c}{WikiQA} & \multicolumn{3}{c}{SNLI} \\ \cline{2-13} 
                                                   & train    & dev     & test   & train     & dev     & test      & train    & dev    & test   & train    & dev   & test  \\ \midrule
\#Q                                                & 9848     & 1958    & 3138   & 13K     & 1K    & 1.8K*2    & 873      & 126    & 243    & 549K   & 9842  & 9824  \\ 
\#C                                                & 5        & 5       & 5      & 50        & 500     & 500       & 10       & 9      & 10     & -        & -     & -     \\ 
\begin{tabular}[c]{@{}c@{}}\#w in P\end{tabular} & 873      & 866     & 914    & -         & -       & -         & -        & -      & -      & -        & -     & -     \\ 
\begin{tabular}[c]{@{}c@{}}\#w in Q\end{tabular} & 10.6     & 10.6    & 10.8   & 7.2       & 7.2     & 7.2       & 6.5      & 6.5    & 6.4    & 14       & 15.2  & 15.2  \\ 
\begin{tabular}[c]{@{}c@{}}\#w in A\end{tabular} & 5.9      & 5.6     & 5.5    & 92.1      & 92.1    & 92.1      & 25.5     & 24.7   & 25.1   & 8.3      & 8.4   & 8.3   \\ \bottomrule
\end{tabular}
\caption{The statistics of different data sets. Q:question/hypothesis, C:candidate answers for each question, A:answer/hypothesis, P:plot, w:word (average).}
\label{table:stat}
\end{table}

In this section, we evaluate our model on four different datasets representing different tasks.
The first three datasets are question answering tasks while the last one is on textual entailment.
The statistics of the four datasets are shown in Table~\ref{table:stat}. 
We will fist introduce the task settings and the way we customize the ``compare-aggregate" structure to each task. 
Then we will show the baselines for the different datasets.
Finally, we discuss the experiment results shown in Table~\ref{table:res}.
%We rank the different comparison functions used in our model in decreasing order of the number of parameters they require.

% we will test our model on four different data sets in different tasks the statistics of which are shown in Table~\ref{table:stat}. Due to the different settings of the task, we adapt our model into task specific structures. The experiment results are shown in Table~\ref{table:res} where our models are ranked according to the number of parameters from the smallest to the largest.

\begin{table}[]
\centering
\label{my-label}
\begin{small}
\begin{tabular}{lccccccccc}
\toprule
\multirow{2}{*}{Models}               & \multicolumn{2}{c}{MovieQA}                          & \multicolumn{3}{c}{InsuranceQA}               & \multicolumn{2}{c}{WikiQA}                & \multicolumn{2}{c}{SNLI}      \\ \cline{2-10} 
                                    & dev                      & test                      & dev           & test1         & test2         & MAP             & \multicolumn{1}{l}{MRR} & train         & test          \\ \midrule
\multicolumn{1}{l}{Cosine Word2Vec} & \multicolumn{1}{l}{46.4} & \multicolumn{1}{l}{45.63} & -             & -             & -             & -               & -                       & -             & -             \\
Cosine TFIDF                        & 47.6                     & \textbf{47.36}            & -             & -             & -             & -               & -                       & -             & -             \\
SSCB TFIDF                         & \textbf{48.5}            & -                         & -             & -             & -             & -               & -                       & -             & -             \\
IR model                   & -                        & -                         & 52.7          & 55.1          & 50.8          & -               & -                       & -             & -             \\
CNN with GESD                       & -                        & -                         & 65.4          & 65.3          & 61.0             & -             & -                  & -             & -             \\
Attentive LSTM                      & -                        & -                         & 68.9          & 69.0          & 64.8          & -               & -                       & -             & -             \\
IARNN-Occam                         & -                        & -                         & 69.1          & 68.9          & \textbf{65.1} & \textbf{0.7341} & \textbf{0.7418}         & -             & -             \\
IARNN-Gate                          & -                        & -                         & \textbf{70.0} & \textbf{70.1} & 62.8          & 0.7258          & 0.7394                  & -             & -             \\
CNN-Cnt                           & -                        & -                         & -             & -             & -             & 0.6520           & 0.6652                  & -             & -             \\
ABCNN                      & -                        & -                         & -             & -             & -             & 0.6921          & 0.7108                  & -             & -             \\
CubeCNN                      & -                        & -                         & -             & -             & -             & 0.7090          & 0.7234                  & -             & -             \\
W-by-W Attention                         & -                        & -                         & -             & -             & -             & -               & -                       & 85.3          & 83.5          \\
match-LSTM                         & -                        & -                         & -             & -             & -             & -               & -                       & 92.0          & 86.1          \\
LSTMN                               & -                        & -                         & -             & -             & -             & -               & -                       & 88.5          & 86.3          \\
Decomp Attention                              & -                        & -                         & -             & -             & -             & -               & -                       & 90.5          & 86.8          \\
EBIM+TreeLSTM                                & -                        & -                         & -             & -             & -             & -               & -                       & 93.0 & \textbf{88.3} \\ \midrule
NN                              & 31.6                     &-                           & 76.8          & 74.9          & 72.4          & 0.7102          & 0.7224                  & 89.3          & 86.3          \\
NTN                          & 31.6                     &-                           & 75.6          & 75.0          & 72.5          & 0.7349          & 0.7456                  &91.6               & 86.3          \\
\textsc{EucCos}                           & 71.9                     &-                           & 70.6          & 70.2          & 67.9          & 0.6740          & 0.6882                  & 87.1          & 84.0          \\ 
\textsc{Sub}                              & 64.9                     &-                           & 70.0          & 71.3          & 68.2          & 0.7019          & 0.7151                  &89.8               & \textbf{86.8} \\
\textsc{Mult}                               & 66.4                     &-                           & 76.0          & 75.2          & \textbf{73.4}          & \textbf{0.7433}          & \textbf{0.7545}                  & 89.7              & 85.8          \\
\textsc{SubMult+NN}                              & \textbf{72.1}            & \textbf{72.9}             & \textbf{77.0}   & \textbf{75.6} & 72.3 & 0.7332 & 0.7477         & 89.4 & \textbf{86.8} \\ 
%weighted sub                         & 30.1                     &-                           & 71.8          & 71.2          & 68.7          & 0.7115          & 0.7251                  &89.7               & 86.0          \\
%weighted mul                         & 30.5                     &-                           & 74.8          & 72.8          & 71.8          & 0.7078          & 0.7196                  &90.3               & 86.1          \\ 
\bottomrule
\end{tabular}
\end{small}
\caption{Experiment Results}
\label{table:res}
\end{table}

\subsection{Task-specific Model Structures}
In all these tasks, we use matrix $\mathbf{Q} \in \mathbb{R}^{d \times Q}$ to represent the question or premise and matrix $\mathbf{A}_k \in \mathbb{R}^{d \times A_k}$ ($k \in [1, K]$) to represent the $k^\text{th}$ answer or the hypothesis.
For the machine comprehension task \textbf{MovieQA}~\citep{MovieQA:cvpr2016}, there is also a matrix $\mathbf{P} \in \mathbb{R}^{d \times P}$ that represents the plot of a movie. 
Here $Q$ is the length of the question or premise, $A_k$ the length of the $k^\text{th}$ answer, and $P$ the length of the plot. 

For the \textbf{SNLI}~\citep{bowman2015large:EMNLP} dataset, the task is text entailment, which identifies the relationship (entailment, contradiction or neutral) between a premise sentence and a hypothesis sentence.
Here $K = 1$, and there are exactly two sequences to match.
The actual model structure is what we have described before.

For the \textbf{InsuranceQA}~\citep{feng2015applying} dataset,
the task is an answer selection task which needs to select the correct answer for a question from a candidate pool. 	
For the \textbf{WikiQA}~\citep{yang2015wikiqa:emnlp} datasets, we need to rank the candidate answers according to a question.
For both tasks, there are $K$ candidate answers for each question.
Let us use $\mathbf{r}_k$ to represent the resulting vector produced by Eqn.~\ref{eqn:aggregate} for the $k^\text{th}$ answer.
In order to select one of the $K$ answers, we first define $\mathbf{R} = [\mathbf{r}_1, \mathbf{r}_2, \ldots, \mathbf{r}_K]$.
We then compute the probability of the $k^\text{th}$ answer to be the correct one as follows:
\begin{eqnarray}
\label{eqn:pqasoft}
p(k | \mathbf{R}) & = & \text{softmax}( \mathbf{w}^\text{T} \tanh(\mathbf{W}^{\text{s}}\mathbf{R} + \mathbf{b}^\text{s} \otimes \mathbf{e}_{K}) + b \otimes \mathbf{e}_{K}),
\end{eqnarray}
where $\mathbf{W}^{\text{s}}\in \mathbb{R}^{l\times nl}$, $\mathbf{w}\in \mathbb{R}^{l}$, $\mathbf{b}^\text{s}\in \mathbb{R}^{l}$, $b\in \mathbb{R}$ are parameters to be learned.

For the machine comprehension task \textbf{MovieQA}, each question is related to Plot Synopses written by fans after watching the movie and each question has five candidate answers.
So for each candidate answer there are three sequences to be matched: the plot $\mathbf{P}$, the question $\mathbf{Q}$ and the answer $\mathbf{A}_k$.
For each $k$, we first match $\mathbf{Q}$ and $\mathbf{P}$ and refer to the matching result at position $j$ as $\mathbf{t}^\text{q}_{j}$, as generated by one of the comparison functions $f$.
Similarly, we also match $\mathbf{A}_k$ with $\mathbf{P}$ and refer to the matching result at position $j$ as $\mathbf{t}^\text{a}_{k, j}$.
We then define
\begin{eqnarray*}
\mathbf{t}_{k, j} & = & 
\begin{bmatrix} 
	\mathbf{t}^\text{q}_j \\
	\mathbf{t}^\text{a}_{k, j}
\end{bmatrix},
\end{eqnarray*}
and
\begin{eqnarray*}
\mathbf{r}_k & = & \text{CNN}([\mathbf{t}_{k, 1}, \ldots, \mathbf{t}_{k, P}]).
\end{eqnarray*}
To select an answer from the $K$ candidate answers, again we use Eqn.~\ref{eqn:pqasoft} to compute the probabilities.

\subsection{Baselines}
Here, we will introduce the baselines for each dataset.
We did not re-implement these models but simply took the reported performance  for the purpose of comparison.

\textbf{SNLI:}
$\bullet$ \textbf{W-by-W Attention}: The model by \cite{rock:ICLR2016}, who first introduced attention mechanism into text entailment. 
$\bullet$ \textbf{match-LSTM}: The model by \cite{wang:NAACL2016}, which concatenates the matched words as the inputs of an LSTM. 
$\bullet$ \textbf{LSTMN}: Long short-term memory-networks proposed by \cite{cheng2016long}. 
$\bullet$ \textbf{Decomp Attention}: Another ``compare-aggregate'' model proposed by \cite{parikh:emnlp2016}. 
$\bullet$ \textbf{EBIM+TreeLSTM}: The state-of-the-art model proposed by \cite{chen2016enhancing:arxiv} on the SNLI dataset.

\textbf{InsuranceQA:}
%\footnote{https://github.com/shuzi/insuranceQA/tree/master/V1}
%InsuranceQA~\citep{feng2015applying} is 
%We randomly select 50 passages, including the correct one, as the candidate pool for training. The candidate pool for the questions in the development and the test datasets are provided by \cite{feng2015applying}. 
$\bullet$ \textbf{IR model}: This model by \cite{bendersky2010:wsdm} learns the concept information to help rank the candidates.
$\bullet$ \textbf{CNN with GESD}: This model by \cite{feng2015applying} uses Euclidean distance
and dot product between sequence representations built through convolutional neural networks to select the answer.
$\bullet$ \textbf{Attentive LSTM}: \cite{tanimproved:acl2016} used soft-attention mechanism to select the most important information from the candidates according to the representation of the questions.
$\bullet$ \textbf{IARNN-Occam}: This model by \cite{wanginner:acl2016} adds regularization on the attention weights.
$\bullet$ \textbf{IARNN-Gate}: This model by \cite{wanginner:acl2016} uses the representation of the question to build the GRU gates for each candidate answer. 

\noindent \textbf{WikiQA:}
% The baselines in this task include: 
$\bullet$ \textbf{IARNN-Occam} and \textbf{IARNN-Gate} as introduced before. 
$\bullet$ \textbf{CNN-Cnt}: This model by \cite{yang2015wikiqa:emnlp} combines sentence representations built by a convolutional neural network with logistic regression.
$\bullet$ \textbf{ABCNN}: This model is Attention-Based Convolutional Neural Network proposed by \cite{yin2015abcnn:tacl}. $\bullet$ \textbf{CubeCNN} proposed by \citet{he:naacl16} builds a CNN on all pairs of word similarity.

\noindent \textbf{MovieQA:}
All the baselines we consider come from \cite{MovieQA:cvpr2016}'s work:
$\bullet$ \textbf{Cosine Word2Vec}: 
%Each sentence in plot/question/answer is represented by the mean-pooling of the pre-trained word embeddings. 
A sliding window is used to select the answer according to the similarities computed through Word2Vec between the sentences in plot and the question/answer.
$\bullet$ \textbf{Cosine TFIDF}: This model is similar to the previous method but uses bag-of-word with tf-idf scores to compute similarity.
$\bullet$ \textbf{SSCB TFIDF}: Instead of using the sliding window method, a convolutional neural network is built on the sentence level similarities.

\subsection{Analysis of Results}

We use accuracy as the evaluation metric for the datasets MovieQA, InsuranceQA and SNLI, as there is only one correct answer or one label for each instance.  
For WikiQA, there may be multiple correct answers, so evaluation metrics we use are
Mean Average Precision (MAP) and Mean Reciprocal Rank (MRR).

We observe the following from the results.
(1) Overall, we can find that our general ``compare-aggregate" structure achieves the best performance on \textbf{MovieQA}, \textbf{InsuranceQA},  \textbf{WikiQA} datasets and very competitive performance on the \textbf{SNLI} dataset. 
Especially for the \textbf{InsuranceQA} dataset, with any comparison function we use, our model can outperform all the previous models. 
(2) The comparison method \textsc{SubMult+NN} is the best in general. 
(3) Some simple comparison functions can achieve better performance than the neural networks or neural tensor network comparison functions. 
For example, the simplest comparison function \textsc{EucCos} achieves nearly the best performance in the \textbf{MovieQA} dataset, and the element-wise comparison functions, which do not need parameters can achieve the best performance on the \textbf{WikiQA} data set.

\subsection{Further Analyses}
%\begin{table}[]
%\small
%\centering
%\begin{tabular}{lcccccc}
%\toprule
%  & NN   & NTN & \textsc{EucCos} & \textsc{Sub} & \textsc{Mult} & \textsc{SubMult+NN} \\ \midrule
%\#parameters in $f$     & 45K  & 3M  & 0      & 0   & 0    & 45K        \\ 
%\#parameters in $f$+CNN & 113K & 3M  & 1K     & 68K & 68K  & 113K       \\ \bottomrule
%\end{tabular}
%\normalsize
%\caption{The number of parameters needed for different comparison functions.}
%\label{my-label}
%\end{table}
\begin{figure}[]
	\centering
	\includegraphics[width=5.5in]{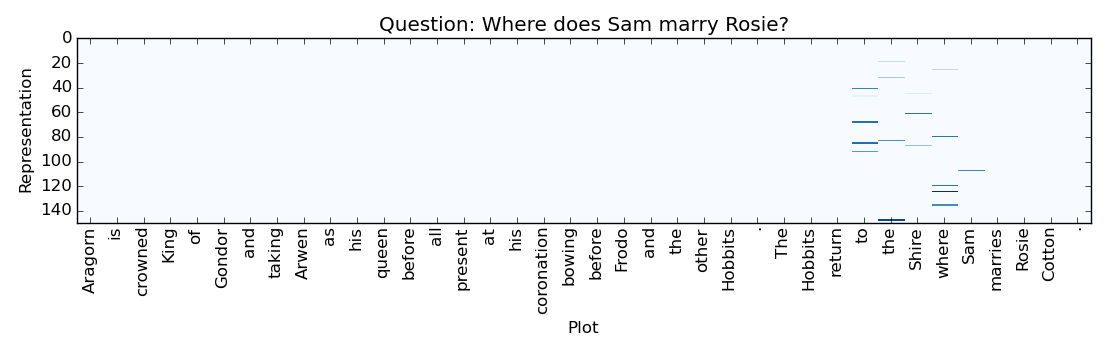}
	\includegraphics[width=5.5in]{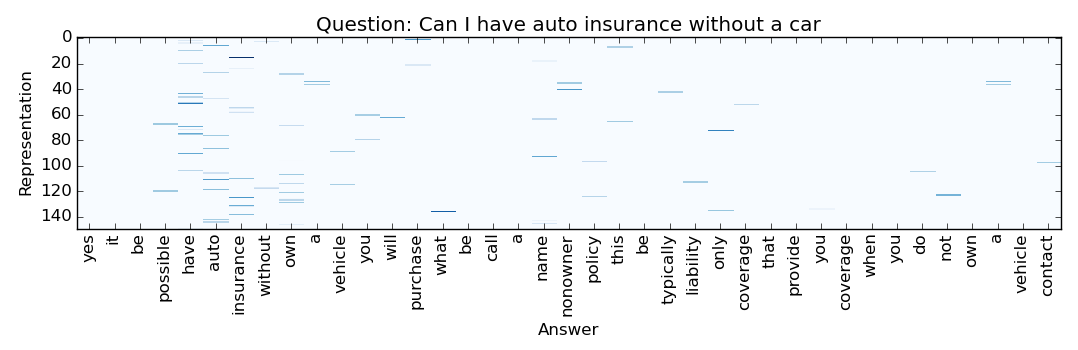}
	\caption{An visualization of the largest value of each dimension in the convolutional layer of CNN. The top figure is an example from the data set \textbf{MovieQA} with CNN window size 5. The bottom figure is an example from the data set \textbf{InsuranceQA} with CNN window size 3.}
	\label{fig:visual}
\end{figure}
To further explain how our model works, we visualize the max values in each dimension of the convolutional layer. 
We use two examples shown in Table~\ref{sample} from MovieQA and InsuranceQA data sets respectively.
In the top of Figure~\ref{fig:visual}, we can see that the plot words that also appear in either the question or the answer will draw more attention by the CNN. 
We hypothesize that if the nearby words in the plot can match both the words in question and the words in one answer, then this answer is more likely to be the correct one. 
Similarly, the bottom one of Figure~\ref{fig:visual} also shows that the CNN will focus more on the matched word representations. 
If the words in one answer continuously match the words in the question, this answer is more likely to be the correct one.

\section{Related Work}

We review related work in three types of general structures for matching sequences.

% With the development of neural networks, the sequences matching information can be represented into a hidden state.
\textbf{Siamense network:} These kinds of models use the same structure, such as RNN or CNN, to build the representations for the sequences separately and then use them for classification.  Then cosine similarity~\citep{feng2015applying,yang2015wikiqa:emnlp}, element-wise operation~\citep{tai2015improved:acl,mou2015:emnlp} or neural network-based combination~\citet{bowman2015large:EMNLP} are used for sequence matching.

\textbf{Attentive network:} Soft-attention mechanism~\citep{bahdanau:ICLR2015} has been widely used for sequence matching in machine comprehension~\citep{hermann:nips2015}, text entailment~\citep{rock:ICLR2016} and question answering~\citep{tanimproved:acl2016}. 
Instead of using the final state of RNN to represent a sequence, these studies use weighted sum of all the states for the sequence representation.

\textbf{Compare-Aggregate network:} This kind of framework is to perform the word level matching~\citep{wang2016machine,parikh:emnlp2016,he:naacl16,trischler:acl2016}. 
Our work is under this framework. 
But our structure is different from previous models and our model can be applied on different tasks. Besides, we analyzed different word-level comparison functions separately.
\section{Conclusions}
In this paper, we systematically analyzed the effectiveness of a ``compare-aggregate" model on four different datasets representing different tasks. 
Moreover, we compared and tested different kinds of word-level comparison functions and found that some element-wise comparison functions can outperform the others.
According to our experiment results, many different tasks can share the same ``compare-aggregate" structure. 
In the future work, we would like to test its effectiveness on multi-task learning.

\bibliography{iclr2017_conference}
\bibliographystyle{iclr2017_conference}

\newpage
\appendix
\section{appendix}
Following are the implementation details. The word embeddings are initialized from GloVe~\citep{pennington2014glove:emnlp2014}.
During training, they are not updated. 
The word embeddings not found in GloVe are initialized with zero.  

The dimensionality $l$ of the hidden layers is set to be 150.
We use ADAMAX~\citep{kingma2014adam:iclr2015} with the coefficients $\beta_1=0.9$ and $\beta_2=0.999$ to optimize the model. 
The batch size is set to be 30 and the learning rate is 0.002.
We do not use L2-regularization. 
The hyper-parameter we tuned is the dropout on the embedding layer. 
For WikiQA, which is relatively small dataset, we also tune the learning rate and batch size. For the convolutional window sizes for MovieQA, InsuranceQA, WikiQA and SNLI, we use [1,3,5], [1,2,3], [1,2,3,4,5] and [1,2,3,4,5], respectively.
\end{document}